\documentclass{article}
\usepackage{spconf,amsmath,graphicx}
\usepackage{xcolor}
\usepackage{mathtools}

\title{When Causal Intervention Meets Adversarial Examples and Image Masking for Deep Neural Networks}
\name{Chao-Han Huck Yang$^{*, 1}$ \thanks{$^{*}$Authors contributed equally to this work}, Yi-Chieh Liu$^{*, 1}$, Pin-Yu Chen$^{2}$, Xiaoli Ma$^{1}$, Yi-Chang James Tsai$^{1}$}
\address{$^{1}$Georgia Institute of Technology, North Ave NW, Atlanta, GA 30332 \\
$^{2}$IBM Research, Yorktown Heights, NY 10598}
\begin{document}
%
\maketitle
\begin{abstract}


Discovering and exploiting the causality in deep neural networks (DNNs) are crucial challenges for understanding and reasoning causal effects (CE) on an explainable visual model. "Intervention" has been widely used for recognizing a causal relation ontologically. In this paper, we propose a causal inference framework for visual reasoning via do-calculus. To study the intervention effects on pixel-level features for causal reasoning, we introduce pixel-wise masking and adversarial perturbation. In our framework, CE is calculated using features in a latent space and perturbed prediction from a DNN-based model. We further provide the first look into the characteristics of discovered CE of adversarially perturbed images generated by gradient-based methods \footnote{~~https://github.com/jjaacckkyy63/Causal-Intervention-AE-wAdvImg}. Experimental results show that CE is a competitive and robust index for understanding DNNs when compared with conventional methods such as class-activation mappings (CAMs) on the Chest X-Ray-14 dataset for human-interpretable feature(s) (e.g., symptom) reasoning. 
Moreover, CE holds promises for detecting adversarial examples as it possesses distinct characteristics in the presence of adversarial perturbations.



\end{abstract}
\section{Introduction}
\label{sec:intro}
In recent years, deep neural networks (DNNs) based models have demonstrated a significant improvement in state-of-the-art results for several vision tasks \cite{yeung2016end, huang2017densely}. Despite their effectiveness, DNNs are notoriously cryptic for their complexity, opacity, and lacking apparent relationship between neural architecture and the function being evaluated by the network. These challenges are amplified especially when DNN models show superior performance over conventional methods (e.g., decision-trees) on human-interpretable  feature learning tasks, such as clinical (i.e., cancer recognition \cite{esteva2017dermatologist}), emotional (i.e., violence detection \cite{sultani2018real}), and social (i.e., behavior reasoning \cite{wang2016deep}) patterns recognition, where even experts did not fully reach a proper consensus. Therefore, interpretability \cite{zhang2018interpretable} and explainable reasoning features of DNNs remain challenging. 
From regulatory aspects (e.g., GDPR \cite{albrecht2016gdpr}), human-interpretable feature selection is crucial for administrative decision-making, as current DNNs highly depend on supervised labels generated by human knowledge but barely explain the learned relationship between its labeled input-feature(s) and predicted output(s). 
The importance of reasoning is also motivated by demystifying the vulnerable properties of DNNs against adversarial examples \cite{goodfellow2014explaining,biggio2018wild,su2018robustness}.

This paper studies the problems of causal interpretability on DNNs for vision tasks. We propose a new visual reasoning method inspired by causal inference embedded DNNs and verify its sensitivity on input masking and adversarial perturbations. We summarize some input intervention-based explanation methods for DNNs as follows.
\begin{figure} 
\begin{center}
   \includegraphics[width=1.0\linewidth]{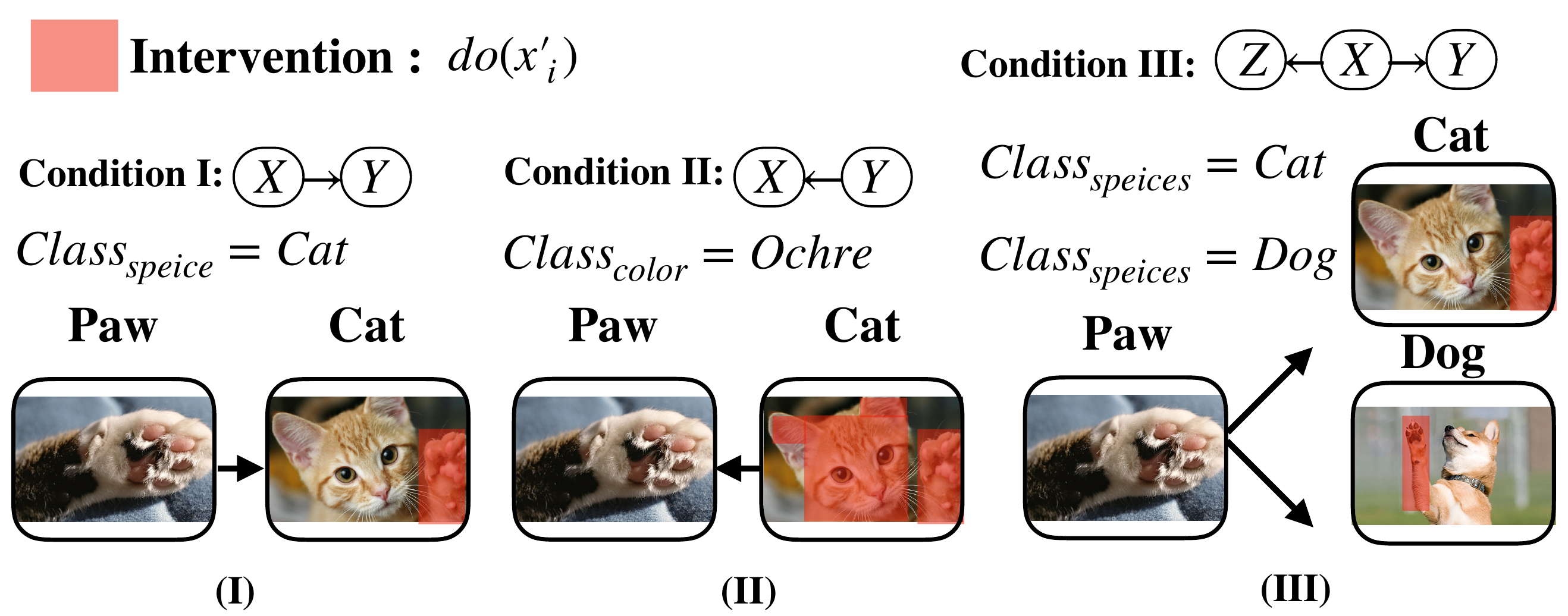}
\end{center}
\vspace{-0.7cm}
   \caption{Visual intervention in three classical causal conditions~\cite{pearl2009causality}: (I) $X \rightarrow{}Y$ for single class counterfactual reasoning; (II) $X \leftarrow{}Y$ for  multi-classes counterfactual reasoning, and (III) $Z \leftarrow{}X\rightarrow{}Y$ for conditional multitasking counterfactual reasoning of a visual perceptual-model. } 
\label{fig:figure1}
\vspace{-0.5cm}
\end{figure}
\\
\textbf{Saliency map based method.} 
Many works employ saliency map for visual explanations.
Class activation mappings \cite{zhou2016learning} (CAMs) have provided visual explanations via tracing gradient-based localization. Gradient information from a penultimate convolutional layer was used in GradCAM \cite{selvaraju2017grad} to identify discriminating patterns in input images. However, CAM-based methods have some limitations on causal reaonsing. They could only conclude that a DNN model would feature with certain localized patterns. Denoting a trained DNNs model as $Y$ and input data as $X$, CAM-based mothods can be viewed as a reverse causation (e.g., $X_{input}\xrightarrow[\overleftarrow{CAM}]{\textbf{DNN Model }}Y_{output}$). Saliency maps \cite{zhou2016learning, selvaraju2017grad} only establish a correlation for interpretability while it is possible to trace a particular image region that is responsible for it to be correctly classified; it cannot elucidate what would happen if a certain portion of the image was masked out. See \cite{pearl2009causality} for a \textbf{What If} question, when a classifier is used in the wild. In addition to this causal reasoning weakness, recent study also shows CAMs barely preserve robustness from a gradient adversarial attacks as a secure reasoning index \cite{prakash2018deflecting}. \\
\textbf{Adversarial robustness.} Although there has recently been a great deal of research in adversarial defenses, many of these methods are later shown to be broken \cite{athalye2018obfuscated}.  The most common, “brute force” solution is adversarial training, where we simply include a mixture of normal and adversarially-generated examples in the training set \cite{kurakin2016adversarial}. One obvious limitation of previous work is the robustness generalizability for  adversarial examples beyond the pre-defined norm or distributional constraints. 
\\
\textbf{Causal reasoning.} Building causal interpretations toward evaluating a robust and interpretable DNNs are still largely unexplored. Causality as a means of explanation has deep roots for factor selection and has received many attentions. The rest of this paper is presented as follows. First, we describe a general framework to build a causal inference to reason over a targeted DNN model via intervention. Second, we discover adversarial robustness between causal factor. Lastly, we propose an adversary-sensitive feature-map to visualize the causal effects of interpreterable features (e.g., symptoms) toward large-scale medical datasets by using DenseNet \cite{huang2017densely}. 
\textbf{Our contributions include:}
\begin{itemize}
    \item The design of a causal inference framework for calculating CE on single task, multi-labels, and multitasking vision tasks. An autoencoder network has been introduced for intervening interpretable feature(s).
    \item The investigation of CE toward highlighting counterfactual intervention by pixel-wise masking and adversarial settings.  
    \item A case study for learning the causal effects from a large-scale medical dataset (Chest X-Ray-14) for reasoning related chest symptoms by Causal Effect Map (CEM). 
\end{itemize}

\section{Related Works}
\label{sec:format}



Causal models have been advocated as a proper method \cite{kusner2017counterfactual} for parameter reasoning in a complex system. Modeling bias and fairness as a counterfactual \cite{kusner2017counterfactual} is a new way to quantify the bias of models. Ideas from causal inference have also been used in providing comprehensive explanations of deep learning models. In \cite{harradon2018causal}, a method is developed to build a causal model over human-centric image-based abstractions of the model, which helps provide the ability to allow users to propose comprehensive queries. With an interpretable concept, the method could learn a causal model relating to network activations and features for classifications. Moreover, as DNNs can be viewed as a graphical model for inference, recent work in \cite{narendra2018explaining} proposed using Structural Causal Models (SCM) to reason over a deep-learning model. The technique includes building the direct acyclic graph (DAG) structure from the DNN, applying a suitable transformation and estimating the causal effect using causal inference.

\section{Visual Causal Intervention}
\label{sec:pagestyle}
\subsection{Causal Theory}
We adopt the widely-accepted Causal Effect (CE) from Pearl's do-calculus\footnote{We use do-operator $do(\cdot)$ as conditional intervention following by the original definition in Pearl 2009~\cite{pearl2009causality}. } for $do(X_i=x_i)\equiv x_i=f_i(pa_i,u_i)$, where $pa_i$ are the parents of variable $X_i$ in a graph, $G$. Selecting two disjoint sets of random variables, $X$ and $Y$, the causal effect of $X$ on $Y$ is denoted as $P(y|do(x))$, which is a function from $X$ to the space of probability distributions on $Y$. As an illustration, here we introduce a graph $G$ including random variables $ \mathbf{X, Y, Z}$  in \textbf{Figure 1} for constructing elemental causal networks for this study.
\\
\textbf{Computing the Effect of Intervention}
An atomic intervention do($X_i = x'_i$) has been used in the previous study \cite{harradon2018causal} as a $truncated~factorization$ for feature selection in a latent space of a targeted DNN system for causal reasoning. Given a joint distribution P($\mathbf{O}, \mathbf{P}, \mathbf{X}$), over a set of DNN with outputs $\mathbf{O}$, inputs $\mathbf{P}$, and intermediate variables $\mathbf{X}$, from the definition in \cite{rosenbaum1983assessing}, measuring CE of an intervention on $x'_i$ on $X_j = x_j$ with all of the evidence $Z$ could be computed as:

\vspace{-0.3cm}
\begin{equation}
Effect(x_i \rightarrow{} x_j, Z)=P(x_j|do(x_{i}'),Z_{X_i})-P(x_j|Z_{X_i})
\label{eq:Eq1}
\end{equation}
The excepted casual effect from \textbf{Eqn. 6} in \cite{harradon2018causal} has been defined as:
\vspace{-0.3cm}
\begin{equation}
E_{X_i}[Effect(x_i \rightarrow{} x_j, Z)]=\sum _{x_i\in X_i}P(X_i=x_i|Z)\times (1)
\label{eq:Eq3}
\end{equation}
Then, we adapt the method in (\ref{eq:Eq3}) on DNN model with (I) single task, (II) multiple tasks, and (III) multiple labels from Figure.\ref{fig:figure1} for computing expected-CE analysis, where $Z_{X_i}$ is a counterfactual setting. 
\\
\textbf{Interventions as Variables} Different from previous approaches to directly measuring the intervened result for calculating CE, alternatively, an intervention $do(X_i = x'_i)$ could be encoded by adding to $G$ a link $F_i \rightarrow{X_i}$, where $F_i$ is a new variable taking values in $\left \{ do(x'_i), idle \right \}$. Here, the notion of $``idle"$ means ``no intervention'' following Pearl's \textbf{Eqn. 3.8} \cite{pearl2009causality}. In this paper, we adapt this concept by introducing masking and adversarial intervention as $F_i$ variables for discovering CE within the DNN in a zero-out evident state, $Z_0$. This conditional probability could be calculated by:

\vspace{-0.15cm}
\begin{equation}
P(x_i|pa'_i) =\left\{\begin{matrix}
 P(x_i|pa_i)& if~F_1 = idle,\\ 
 0& if~F_i = do(x'_i)~and~x_i\neq x'_i,  \\ 
 1& if~F_i = do(x'_i)~and~x_i = x'_i.
\end{matrix}\right.
\label{eq:Eq2}
\end{equation}
The advantages of this augmented causal graph in (\ref{eq:Eq2}) are its capability for showing the ramification of spontaneous change in $f_i$ from $F_i$ instead of merely replacing $f_i$ by a constant. 


\begin{figure}[h]
\begin{center}
\vspace{-2mm}
   \includegraphics[width=1.0\linewidth]{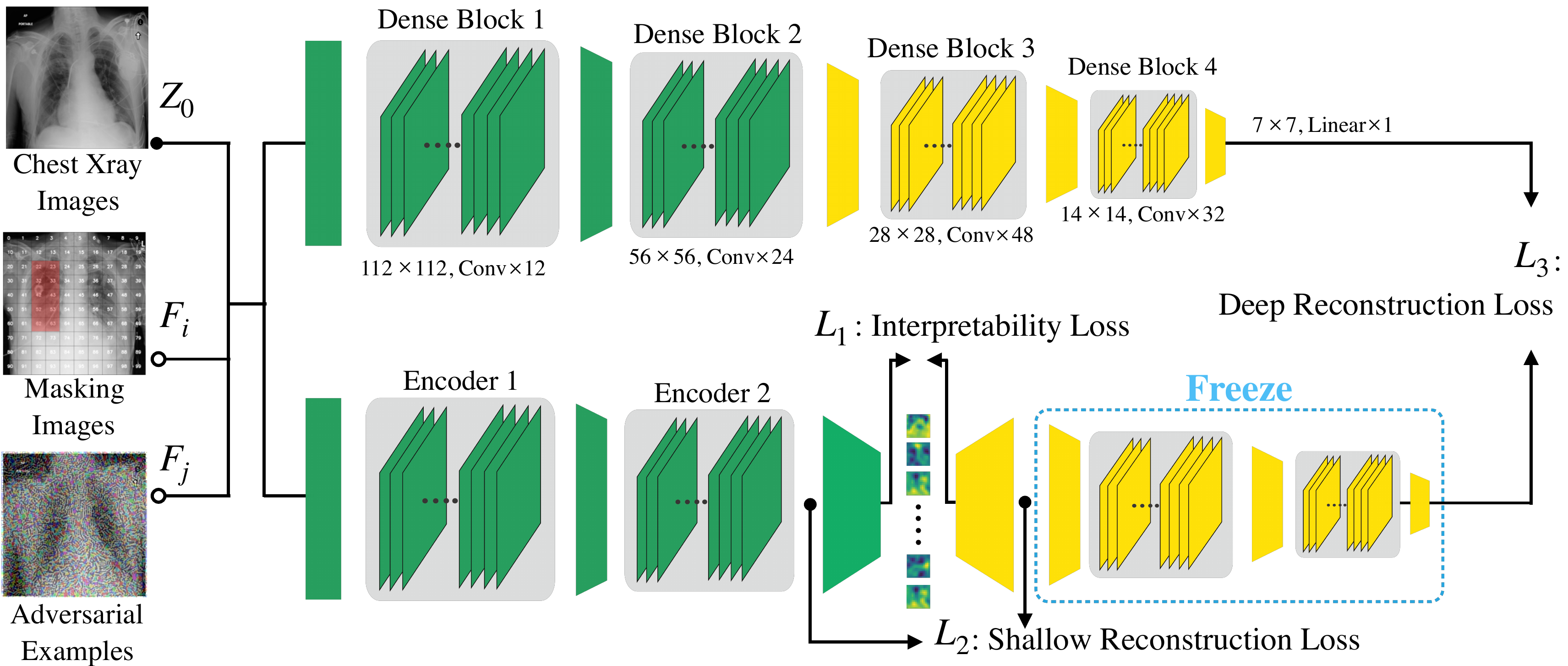}
\end{center}
\vspace{-0.7cm}
   \caption{Framework and deep network topology. We use pixel-wise masking as $F_i$ and adversarial examples as $F_j$ intervention variables in the $Z_0$ (zero-out) counterfactual setting. } 
\label{fig:figure2}
\vspace{-0.6cm}
\end{figure}
\subsection{Concept extraction by Autoencoder}
\textbf{Autoencoder networks} Our autoencoder consists of five convolutional layers. The encoder and decoder each have two symmetrical layers, and they aim to capture the textural features corresponding to human visual perception. However, our decoder reconstructs slightly altered network activations while ensuring significant downstream network activations are unaffected. We mainly designed a low-dimensional layer in the middle to integrate scattered features into concepts. This is inspired by how our brain pays attention, processes information and interpretsthe perceived inputs. 
\\
\textbf{Build up a Causal Baseline} To attain better causality measurement, we trained our model with a designed loss function composed of three parts from the existing studies \cite{harradon2018causal} as casual baseline: interpretability loss, shallow reconstruction loss, and deep reconstruction loss. Shallow reconstruction loss is simply the $L_1$ norm of the difference between the input and output of autoencoder to represent the activations of the network. Also, we applied the deep reconstruction loss in the form of the KL-divergence between the output probability distribution of original and autoencoder-inserted network. Our total loss function for the causality model is (from Harradon et al. 2018 \textbf{Eqn. 7} in \cite{harradon2018causal}):
\\
$L(\theta; x_i) = \lambda_{shallow}\times L_{shallow}(\theta; x_i) +
\lambda_{deep}\times L_{deep}(\theta; x_i)
\\
\indent\indent\indent+\lambda_{interpretability}\times L_{interpretability}(\theta; x_i)$ (3)
\\
Our method is based on the proposed loss in the previous work \cite{harradon2018causal} to reproduce the measurement of causality.


\subsection{Intervention Methods}
In the zero-out ($Z_0$) setting, we discuss two types of perturbation as independent intervention variables for computing expected-CE: (1) pixel-wise masking (PWM) and (2) adversarial noise. The purpose of pixel-wise perturbation, $F_i$, is to recognize the causal relationship between a local pixel and the final output result of a DNN. We  apply the following adversarial noises on images as
another intervention $F_j$. \\
\textbf{Fast Gradient Sign Method (FGSM)}
Fast Gradient Sign Method (FGSM) \cite{goodfellow2015laceyella} is a  gradient-based adversarial noise to generate adversarial  examples by one step gradient update along the direction of the sign of gradient at each pixel.
\\
\textbf{Jacobian-Based Saliency Map (JBSM) Attack}
Papernot et al. \cite{papernot2016distillation} designed an efficient saliency adversarial map, called Jacobian-based Saliency  Map Attack by saturating a few pixels in a given image to their maximum or minimum values. 
\\
\textbf{BIM and PGD Attacks}
Kurakin et al. \cite{kurakin2016adversarial} applied extended Fast Gradient Sign method by running a finer optimization (smaller change) for multiple iterations Basic Iterative Method (BIM) attack. Projected gradient descent (PGD) attack has been  introduced by subjecting to a constraint \cite{kurakin2016adversarial}. 
\section{Experiments}
\label{sec:typestyle}

We use three benchmark image datasets which have been proved robust to reassure that the causal effect would not be compromised by the defect of data. The datasets are fashion-MNIST, ImageNet and NIH-ChestX-Ray-14. We investigate $X \rightarrow{}Y$ for single feature causal intervention. For adversarial setting in our experiments, we use maximum $L_\infty$ perturabtion strength $\epsilon$ = $0.3$ and iterative steps = 10.\\
\textbf{Numerical Validation on Bird Dataset}
In previous work, the VGG19 \cite{simonyan2014very} model applied to bird200 \cite{branson2010visual} had significant causal effect with intervention at layer 18 feature 2. In our work, we also use VGG19 to reproduce the procedure and test the causality by intervening the same position. The expect causal effect is $7.6341 \times 10^{-3}$, which suffices to validate our method and proved the potential of better interpretability. 
\subsection{Quantitative Study on NIH-ChestX-Ray-14}
\begin{figure}[h]
\begin{center}
   \includegraphics[width=1.0\linewidth]{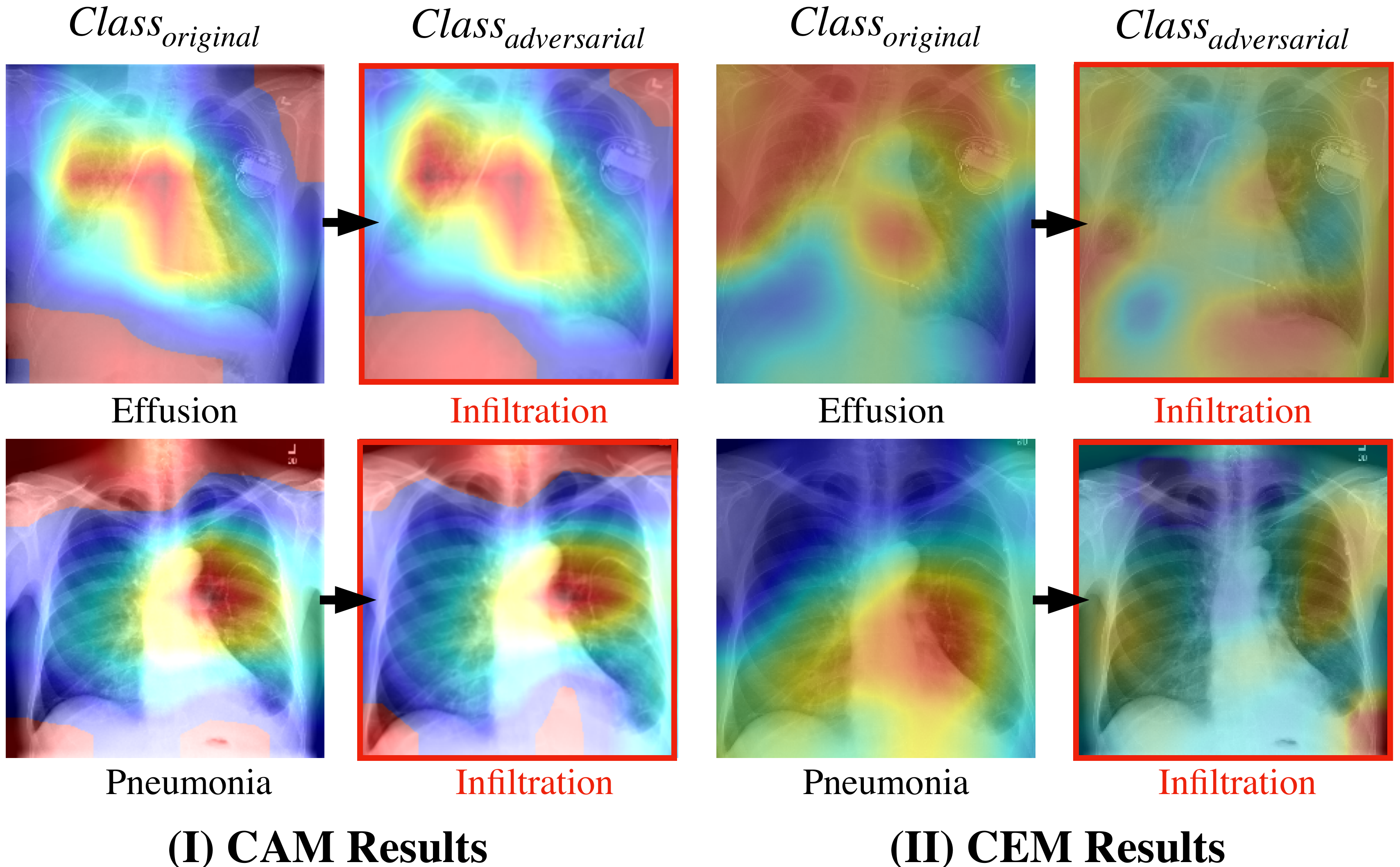}
\end{center}
\vspace{-0.7cm}
   \caption{Sensitivity for adversarial setting between (I) Class Activation Map (CAM) \cite{zhou2016learning} and (II) Causal Effect Map (CEM) on ChestX-ray 14 for clinical symptoms labels \cite{rajpurkar2017chexnet}. } 
\label{fig:figure3}
\end{figure}

NIH-ChestX-Ray-14 \cite{wang2017chestxray} is the largest available chest Xray dataset. Despite of anatomical obstacles, CheXNet model could attain competitive classification performance with radiologists \cite{rajpurkar2017chexnet}. Different from various classes in ImageNet, the interpretable patterns in ChestX-ray14 are  clinical symptoms, which could only be recognized by radiologists and hardly identified by CAM against adversarial setting in Figure \ref{fig:figure3}.\\      
\textbf{CheXNet}
We implemented CheXNet \cite{rajpurkar2017chexnet} as a 121-layer convolutional neural network trained on ChestX-ray14 dataset contains frontal-view chest X-ray images and labeled with up to 14 different thoracic diseases. We modified the DenseNet121 model by replacing the final fully connected layer and apply a sigmoid layer. We use Adam with 0.001 learning rate, $\beta1$ as 0.9, $\beta 2$ as 0.999. We choose binary cross entropy, and the classifier ended up achieved 85.08\% accuracy. Besides, we investigate $Y \rightarrow{}X$ by masking the known symptoms and generate negative Expected-CE as ${-1.624}\times 10^{-4}$, which provides inference. For  $Z \leftarrow{}X \rightarrow{}Y$, we zero-out the nodes which activate additional feature identified as a pulmonary disease as an additional label suggested by a pulmonologist. However, CE of this multitasking reasoning is not obvious.
\textbf{Intervention by pixel-wise Masking}
For PWM, we intervene with the visual traits of inputs. We mask 10 \% area of the images randomly to explore
the impact of masking on the causal effect. We could assure the intervention without bias with a chance to observe the region of interest of the causal model. The results in Table 1, 3, and 4 show the value of Expected-CE deceases by PWM, which indicate a declined causal importance inferences by pixel-wise perturbation.
\\
\textbf{Intervention by Adversarial Attacks}
From Table 2., we observe a numerical difference of Expected-CE by four gradient-based adversarial attacks, where FGSM has a larger negative causal effect with \textbf{-5.6129 $\times 10^{-6}$}. We then adapt FGSM to evaluate generic datasets in Section 4.2. 

The presence of malicious adversarial-noise via CAMs such as (I) in Fig. \ref{fig:figure3} is often hardly to detect if the dataset has similar visual patterns (e.g., clinical patterns). We propose an adversary-sensitive causal effect map (CEM) by activated weights of CE in \textbf{Eqn. 1} on each pixel toward classifying a specific class.

\vspace{-0.4cm}
\begin{table}[h]
\caption {Expected-CE on CheXNet} \label{tab:table1}
\begin{tabular}{|c|c|l|}
\hline
Level(L), Node(N)                 & $Z_0$                        & $F_i$ = PWM \\ \hline
3,4                        & 4.5076$\times 10^{-3}$                       & 7.2356$\times 10^{-7}$     \\ \hline
6,5                        & 2.843$\times 10^{-3}$                        & 1.2154$\times 10^{-5}$     \\ \hline
6,10                       & 3.1939$\times 10^{-3}$                       & 9.0066$\times 10^{-6}$     \\ \hline
8,5  & 3.1939$\times 10^{-3}$  & 1.1536$\times 10^{-5}$     \\ \hline
10,7 & 1.3775$\times 10^{-2}$  & -1.1506$\times 10^{-5}$    \\ \hline
\end{tabular}
\end{table}
\vspace{-0.6cm} 
\begin{table}[h]
\caption {CheXNet ($F_j$ on L = 10, N = 7)} \label{tab:table2}
\begin{tabular}{|c|c|}
\hline
$F_j$ = Types of Adversarial Attack                    & Expected-CE                    \\ \hline
FSGM                      & \textbf{-5.6129$\times 10^{-6}$}                   \\ \hline
BIM                       & 4.3435$\times 10^{-5}$                       \\ \hline
JBSM         & 7.7548$\times 10^{-5}$                       \\ \hline
{PGD} & {-3.9605$\times 10^{-6}$} \\ \hline
\end{tabular}
\end{table}

\vspace{-0.3cm}
\subsection{Other Image Classification Datasets}
\textbf{Fashion-MNIST}
The fashion-MNIST dataset comprises of 28$\times$28 grayscale images of 70,000 fashion products from 10 categories, with 7,000 images per category. \cite{2017arXiv170807747X}. We use shallow model consists of 2 convolutional layers. The weights are trained using Adam with initial learning rate 0.001. We train the model using minibatches of size 100. The classifier with lowest validation loss has 89.6\% accuracy.
\begin{table}[h]
\caption {Expected-CE on Fashion-MNIST}\label{tab:table3}
\begin{tabular}{|c|c|c|c|}
\hline
L, N & $Z_0$      & $F_i$ = PWM     & $F_j$ = FGSM \\ \hline
2,3 & 4.9543$\times 10^{-4}$ & -1.8947$\times 10^{-5}$ & -8.0042$\times 10^{-6}$       \\ \hline
2,7 & 2.4600$\times 10^{-4}$ & -4.1068$\times 10^{-5}$ & -2.5742$\times 10^{-5}$       \\ \hline
\end{tabular}
\end{table}
\\
\textbf{ImageNet}
The ImageNet \cite{deng2009imagenet} 2012 classification dataset consists 1.2 million images for training, and 50,000 for validation, from 1, 000 classes. As a benchmark dataset, we use it to demonstrate visual causality. We use VGG16 model which has 16 convolutional layers and is pretrained by ImageNet with 86\% top-1 accuracy.
\vspace{-0.3cm}
\begin{table}[h]
\caption {Expected-CE on ImageNet}\label{tab:table4}
\begin{center}
\begin{tabular}{|c|c|c|c|}
\hline
L, N & $Z_0$ & $F_i$ = PWM     & $F_j$ = FGSM \\ \hline
3,2         & 4.2027$\times 10^{-4}$    & 6.0993$\times 10^{-5}$     & -5.1834$\times 10^{-5}$       \\ \hline
9,10        & 8.9322$\times 10^{-3}$    & 7.2438$\times 10^{-5}$     & - 2.8862$\times 10^{-4}$    \\ \hline
13,6        & 6.7643$\times 10^{-4}$    & 2.3943$\times 10^{-6}$     & -6.8749$\times 10^{-5}$       \\ \hline
\end{tabular}
\end{center}
\vspace{-0.3cm}
\end{table}
\subsection{Evaluation of Causal Effect}
To validate the reliability of our approach, we also implemented our causal inference framework on Fashion-MNIST and ImageNet. We applied random pixel-wise feature masking and adversarial noise separately to the input images with $Z_0$ coded feature map. Table 3 and 4 show that comparing to $Z_0$, Expected-CE of these two datasets decreased drastically once we applied random masking and adversarial noise, which are similar to the results of CheXNet. The results suggest that our causality framework is generic and can be applied to different datasets (please refer our reproducible code.) 


\section{Conclusion}
\label{sec:majhead}
With our proposed framework, experimental results on Fashion-MNIST, ImageNet, and ChestX-ray14 datasets show that CE is a competitive and robust index for understanding DNNs for generic and human-interpretable feature(s) reasoning. CE also shows high sensitivity in the presence of adversarial perturbations and could be used to identify adversarial examples. 
{ \small
\bibliographystyle{IEEEbib}
\bibliography{strings,refs}
}
\end{document}